\documentclass[10pt,twocolumn,letterpaper]{article}

\usepackage{cvpr}              

\usepackage{colortbl}
\usepackage{booktabs}
\usepackage{makecell}
\usepackage{multirow}
\usepackage{array}
\usepackage{siunitx}
\usepackage{caption}
\usepackage{enumitem}

\definecolor{cvprblue}{rgb}{0.21,0.49,0.74}
\usepackage[pagebackref,breaklinks,colorlinks,allcolors=cvprblue]{hyperref}

\def\paperID{*****} 
\def\confName{CVPR}
\def\confYear{2025}

\title{Spatially-Aware Evaluation of Segmentation Uncertainty}

\author{
  Tal Zeevi \quad
  Eléonore V. Lieffrig \quad
  Lawrence H. Staib\thanks{These authors contributed equally to this work.} \quad
  John A. Onofrey\footnotemark[1]
  \vspace{5pt}
  \\
  Yale University
}

\makeatletter
\def\@maketitle{%
  \vspace*{-4em} 
  \begin{center}
    {\small\color{gray}
Presented at the 4th Workshop on Uncertainty Quantification for Computer Vision,
held in conjunction with the \\ IEEE/CVF Conference on Computer Vision and Pattern Recognition (CVPR) on June 11, 2025 in Nashville, Tennessee.\\
This version is not included in the official proceedings.\par}
    \vskip 4em
    {\Large \bfseries \@title \par}
    \vskip 1.5em
    {\large  \@author \par}
    \vskip 1.5em
  \end{center}
}
\makeatother

\begin{document}
\twocolumn[{%
\renewcommand\twocolumn[1][]{#1}%
\maketitle
\begin{center}
    \centering
    \captionsetup{type=figure}
    \includegraphics[width=0.9\textwidth]{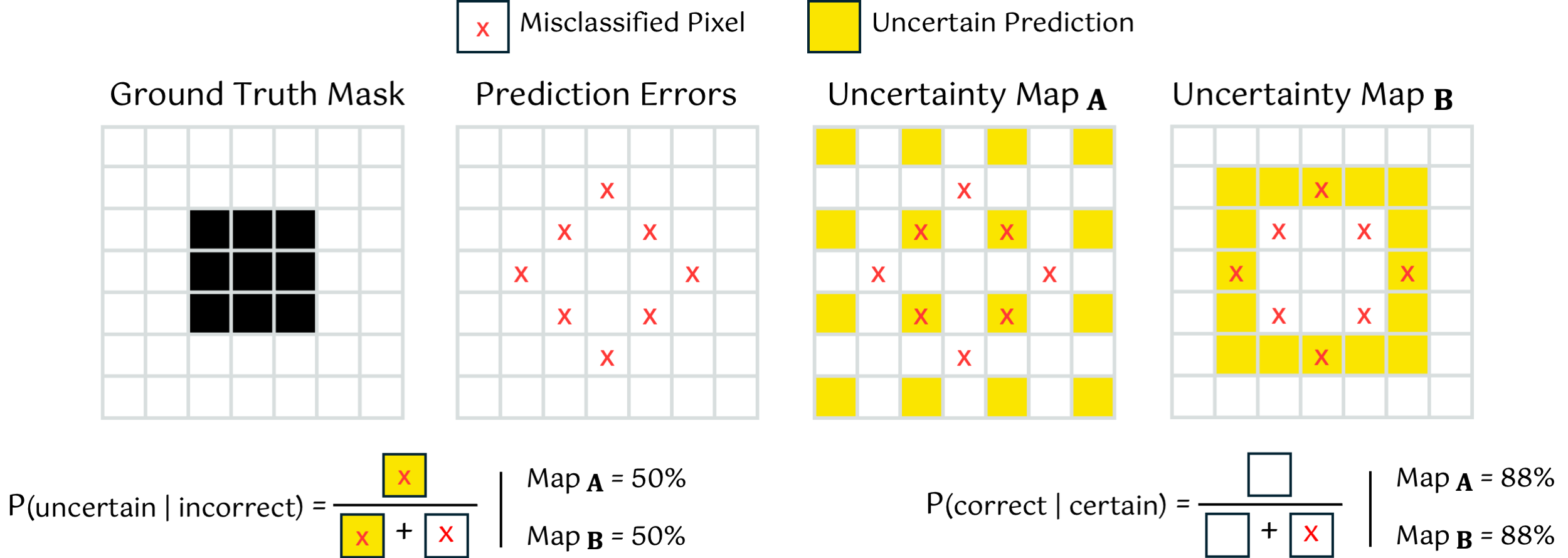}
    
    \captionof{figure}{\textbf{Common metrics for evaluating segmentation uncertainty fail to capture important spatial patterns.} 
    Two uncertainty maps (A and B) receive identical performance metrics despite fundamentally different spatial distributions. Map A shows uniformly scattered uncertainty while Map B concentrates it around anatomical boundaries—a critical distinction in medical image segmentation where boundary ambiguity is clinically expected.}
    \label{fig:uncertainty_maps}
    \vspace{10pt}
\end{center}%
}]

    

\maketitle
\renewcommand{\thefootnote}{\fnsymbol{footnote}}
\footnotetext[1]{These authors contributed equally to this work.}

\begin{abstract}

Uncertainty maps highlight unreliable regions in segmentation predictions. However, most uncertainty evaluation metrics treat voxels independently, ignoring spatial context and anatomical structure. As a result, they may assign identical scores to qualitatively distinct patterns (e.g., scattered vs. boundary-aligned uncertainty). We propose three spatially aware metrics that incorporate structural and boundary information and conduct a thorough validation on medical imaging data from the prostate zonal segmentation challenge within the Medical Segmentation Decathlon. Our results demonstrate improved alignment with clinically important factors and better discrimination between meaningful and spurious uncertainty patterns.

\end{abstract}
  
\section{Introduction}
\label{sec:intro}

Uncertainty quantification is essential for assessing the reliability of segmentation models, especially in high-stakes domains such as medical imaging and autonomous driving. In these settings, boundary uncertainty can have serious consequences, impacting treatment planning or compromising safety. While many methods generate uncertainty maps, their evaluation typically adapts classification-based metrics to voxel-level segmentation uncertainty~\cite{mukhoti2018evaluating, laves2020well, mobiny2021dropconnect}, ignoring spatial coherence and anatomical boundaries.

This can obscure clinically important distinctions, as in medical imaging where uncertainty often appears along ambiguous anatomical boundaries, or lead to safety-critical failures in other domains. Existing metrics fail to capture these spatial differences, treating uncertainty as independent at each voxel rather than considering its alignment with anatomical structures (\cref{fig:uncertainty_maps}).

To address this limitation, we propose three spatially aware evaluation metrics that explicitly incorporate boundary information and spatial correlations: (i) Boundary Uncertainty Concentration (BUC) quantifies whether uncertainty is localized at predicted boundaries; (ii) Boundary-Aware Expected Calibration Error (BA-ECE) extends calibration error analysis by binning uncertainty scores based on distance to the ground-truth boundary; and (iii) Spatially-Aware Calibration Error (SPACE) evaluates local alignment between uncertainty and actual errors using spatial smoothing. We validate these metrics on the prostate zonal segmentation challenge from the Medical Segmentation Decathlon \cite{antonelli2022medical}, demonstrating that they better distinguish clinically meaningful uncertainty from spurious noise, complementing existing voxel-wise approaches.

\section{Related Work: Voxel-wise Metrics}

\begin{itemize}

\item \textbf{Reliability metrics.} 
Early work in uncertainty evaluation borrows from classification calibration, where metrics like the \emph{Expected Calibration Error} (ECE)~\cite{guo2017calibration}, \emph{Maximum Calibration Error} (MCE)~\cite{naeini2015obtaining}, and variants (e.g., SCE, ACE, TACE)~\cite{nixon2019measuring} compare predicted confidence to observed error rates. When applied to segmentation tasks, each voxel is treated as an independent instance of classification, with the calibration error computed by binning voxel-level predictions. The \emph{Expected Uncertainty Calibration Error} (UCE)~\cite{laves2020well} follows the same formulation as ECE but applies it to uncertainty scores instead of confidence. While these metrics provide a compact summary of reliability, they remain spatially agnostic, ignoring spatial dependencies among voxels.
\\
\item \textbf{Discrimination metrics.}
Another approach consists in viewing uncertainty prediction as a binary classifier detecting segmentation errors vs.\ correct predictions. Standard classification metrics, such as \emph{Area Under the ROC Curve} (AUC-ROC) and \emph{Area Under the Precision-Recall Curve} (AUC-PR)~\cite{hendrycks2016baseline, zeevi2024monte}, measure how well high-uncertainty regions align with actual errors. Threshold-based methods, such as \emph{Accuracy versus Uncertainty} (AvU)\cite{krishnan2020improving} and \emph{Patch Accuracy versus Patch Uncertainty} (PAvPU)\cite{mukhoti2018evaluating}, evaluate the discrimination accuracy of uncertainty scores, with the latter applied to local patches. Although effective at ranking uncertain voxels, these metrics are not inherently spatial: they do not penalize scattered noise more than coherent boundary uncertainty, as long as both sets of voxels are similarly predictive of errors.
\\
\item \textbf{Selective prediction metrics.}
In practical applications, one may \emph{reject} or \emph{defer} uncertain predictions. \emph{Area Under the Risk-Coverage Curve} (AURC)~\cite{geifman2018bias} measures how well the model reduces error by discarding highly uncertain voxels, and \emph{AU-ARC}~\cite{nadeem2009accuracy} similarly tracks accuracy vs.\ fraction rejected. While useful for system-level triage, these also remain largely insensitive to spatial distribution. A model may discard 5\% of all voxels, yet remain blind to localized boundary ambiguities.

\end{itemize}

\noindent In summary, voxel-wise metrics offer valuable information on calibration and discrimination but do not differentiate boundary misalignment from diffuse mispredictions. Our work complements these approaches by explicitly modeling spatial structure and boundary proximity.

\section{Proposed Spatially-Aware Metrics}
We introduce three metrics that integrate spatial structure into uncertainty evaluation. Let:

\begin{itemize}
\item \(\upsilon(x)\) denote the \emph{prediction uncertainty} at a given voxel  \(x\). Larger \(\upsilon(x)\) indicates lower model confidence.
\item \(\epsilon(x)\) denote the \emph{binary error indicator}, i.e., \(\epsilon(x) = 1\) if the model's segmentation is incorrect at \(x\), and 0 otherwise.
\end{itemize}

\subsection{Boundary Uncertainty Concentration (BUC)}
In medical image segmentation, errors often occur near object boundaries, where delineation is inherently ambiguous. Let $R$ be the set of voxels within a chosen distance to the predicted boundary (e.g., the 95th percentile Hausdorff distance). We define the mean uncertainty inside and outside the boundary region $R$ as:

\[
\mu_\upsilon^R
= 
\frac{1}{|R|}
\sum_{x \,\in\, R} 
\upsilon(x)
\quad \text{and} \quad
\mu_\upsilon^{\overline{R}}
= 
\frac{1}{|\overline{R}|}
\sum_{x \,\notin\, R} 
\upsilon(x),
\]
The \emph{boundary uncertainty concentration} is then defined as:

\[
\mathrm{BUC} 
=
\frac{\mu_\upsilon^R}
     {\mu_\upsilon^R 
     + \mu_\upsilon^{\overline{R}}}.
\]
A BUC value close to 1 indicates that high prediction uncertainty is mostly localized along the boundary.

\subsection{Boundary-Aware ECE (BA-ECE)}
\label{subsec:baece}
Standard calibration metrics bin voxels by confidence level; here, we instead group them into \emph{distance bands} based on proximity to the \emph{ground-truth} boundary. For each voxel \(x\), let \(d(x)\) denote its shortest distance to the boundary. We partition distances into \(K\) bands \(\{b_1, \dots, b_K\}\), where each band is defined as:

\[
b_i = \{x \mid d(x) \in \Delta_i\}
\]

\noindent with \(\Delta_i\) representing the \(i\)-th distance range. For each band \(b_i\), we compute the mean predicted uncertainty and mean observed error:

\[
\mu_\upsilon^{b_i} = \frac{1}{|b_i|} \sum_{x \,\in\, b_i} \upsilon(x),
\quad
\mu_\epsilon^{b_i} = \frac{1}{|b_i|} \sum_{x \,\in\, b_i} \epsilon(x).
\]

We then define the \emph{boundary-aware expected calibration error} as:

\[
\mathrm{BA\text{-}ECE}
=
\sum_{i=1}^K 
w_i
\Bigl\lvert
\mu_\upsilon^{b_i}
\;-\;
\mu_\epsilon^{b_i}
\Bigr\rvert.
\]

\noindent where \(w_i\) is a distance-based weight inversely proportional to the average distance of voxels in band \(b_i\) from the boundary. Discrepancies in near-boundary
bands reflect miscalibration in critical regions.

\subsection{Spatially-Aware Calibration Error (SPACE)}
SPACE measures the local spatial alignment between a predicted uncertainty map $U$ and the binary error map $E$. Convolving both with a Gaussian kernel $G_{\sigma}$ gives a spatially smoothed comparison:
\[
\mathrm{SPACE} = \mathrm{mean}\Bigl\lvert \bigl(G_{\sigma} * U\bigr) - \bigl(G_{\sigma} * E\bigr)\Bigr\rvert.
\]
Lower values indicate that uncertainty scores closely track actual errors within local neighborhoods (determined by $\sigma$).

\begin{figure*}[t]
    \centering

    \begin{subfigure}[t]{0.31\textwidth}
        \centering
        \includegraphics[width=\linewidth]{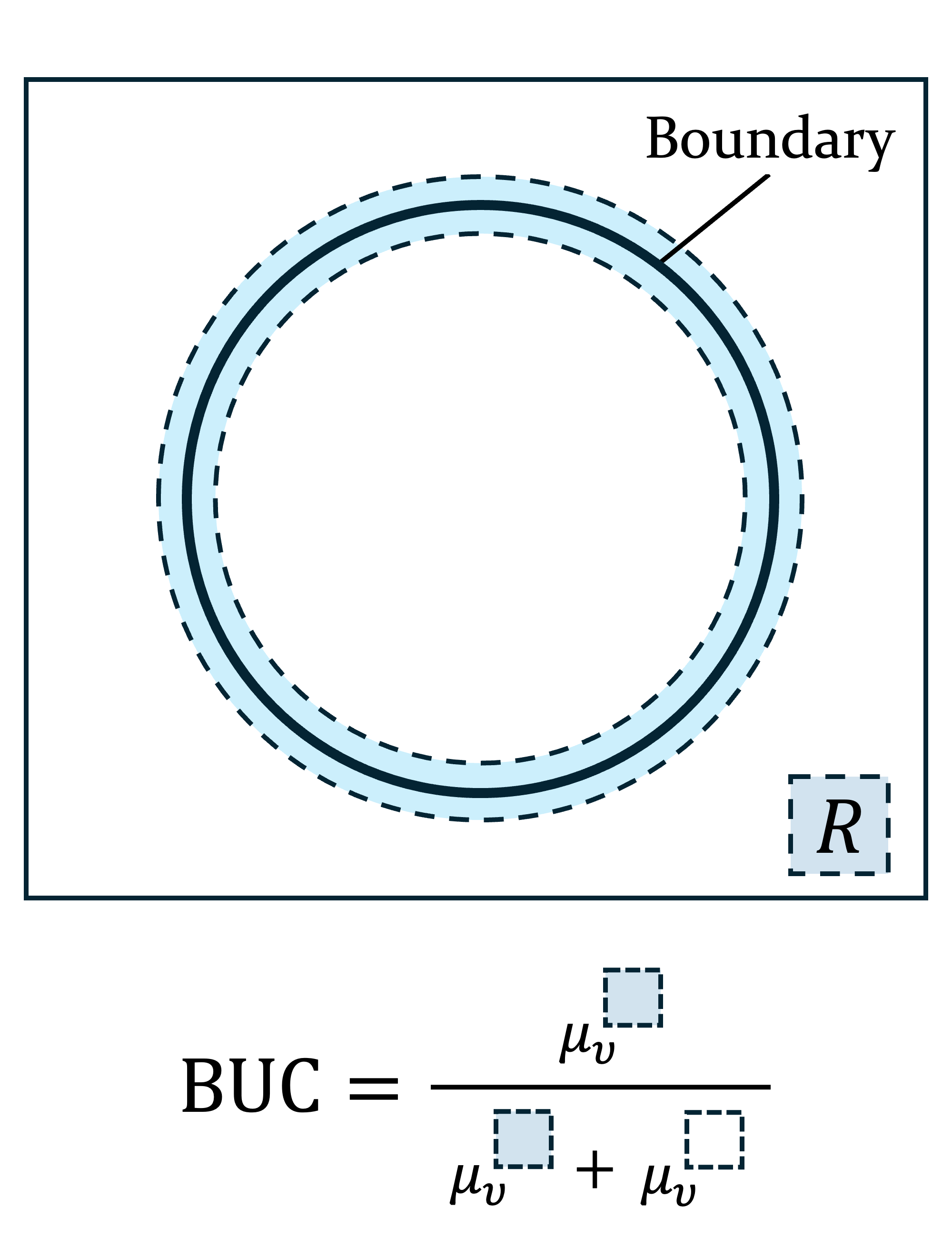}
        \caption{Boundary Uncertainty Concentration \textbf{(BUC)}}
        \label{fig:buc}
    \end{subfigure}
    \hfill
    \begin{subfigure}[t]{0.31\textwidth}
        \centering
        \includegraphics[width=\linewidth]{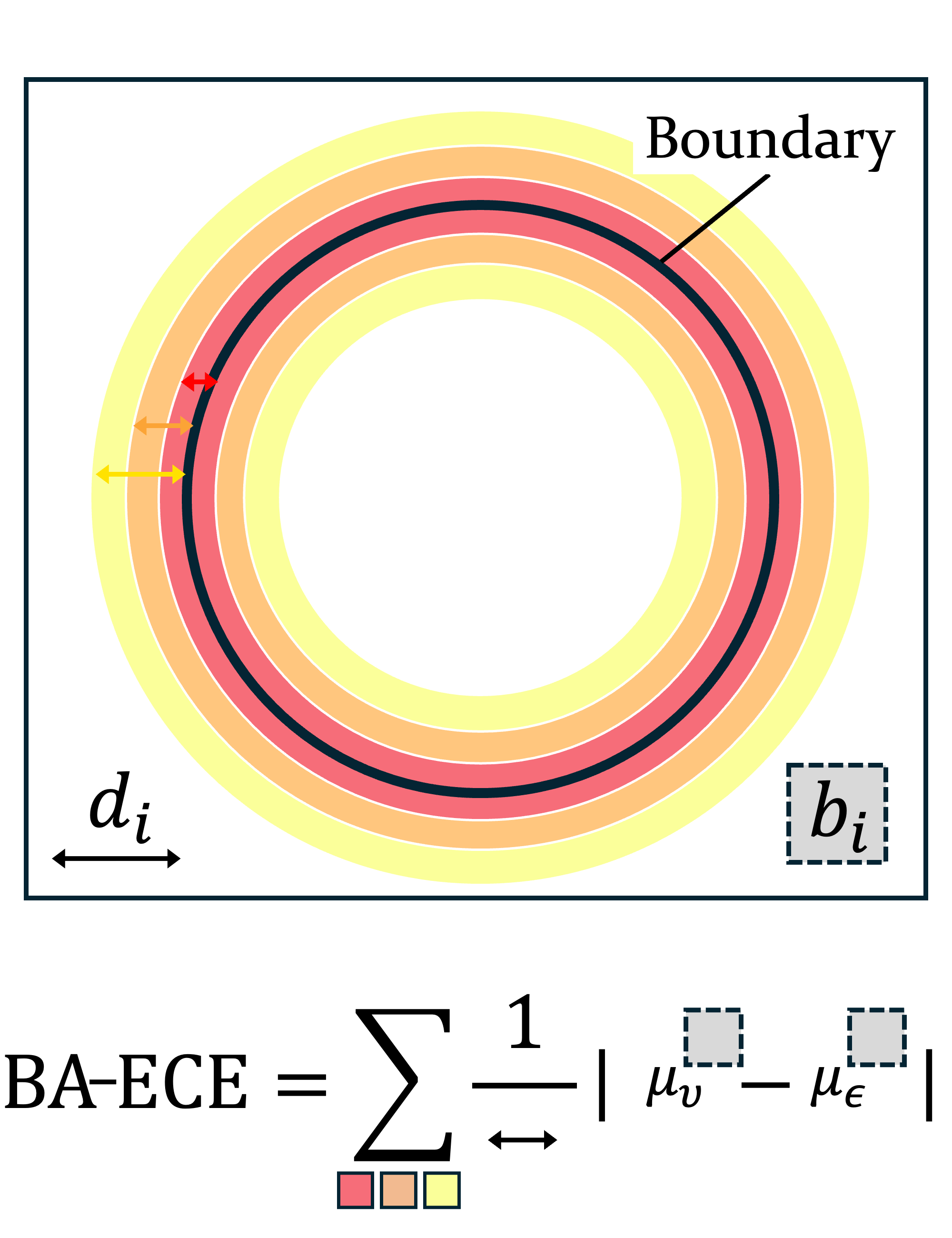}
        \caption{Boundary-Aware ECE \textbf{(BA-ECE)}}
        \label{fig:baece}
    \end{subfigure}
    \hfill
    \begin{subfigure}[t]{0.31\textwidth}
        \centering
        \includegraphics[width=\linewidth]{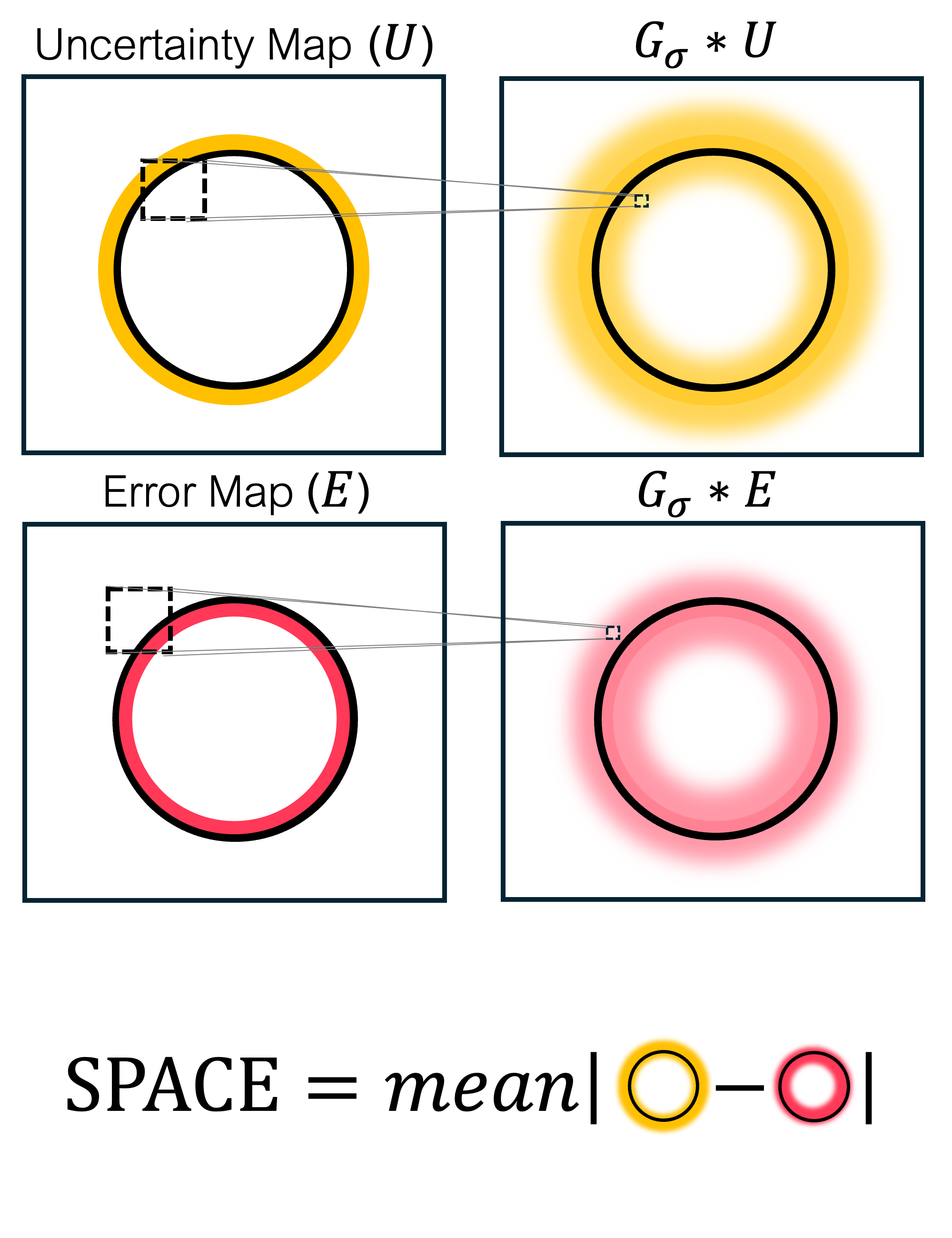}
        \caption{Spatially-Aware Calibration \textbf{(SPACE)}}
        \label{fig:space}
    \end{subfigure}

    \caption{\textbf{Schematic illustration of the proposed spatially-aware uncertainty metrics.} \textbf{(a)} Boundary Uncertainty Concentration \textbf{(BUC)}; The light blue region highlights areas near the object boundary where uncertainty is expected to be higher. \textbf{(b)} Boundary-Aware Expected Calibration Error \textbf{(BA-ECE)}; Voxels are partitioned into distance-based bands relative to the ground-truth boundary to assess spatially resolved calibration. \textbf{(c)} Spatially-Aware Calibration Error \textbf{(SPACE)}; Top-left: high uncertainty near but outside the boundary. Bottom-left: errors within the boundary. Right: smoothing reveals local overlap, capturing spatial correspondence.}
    \label{fig:combined_metrics}
\end{figure*}
\begin{figure*}[ht]
    \centering
    \includegraphics[width=1.00\textwidth]{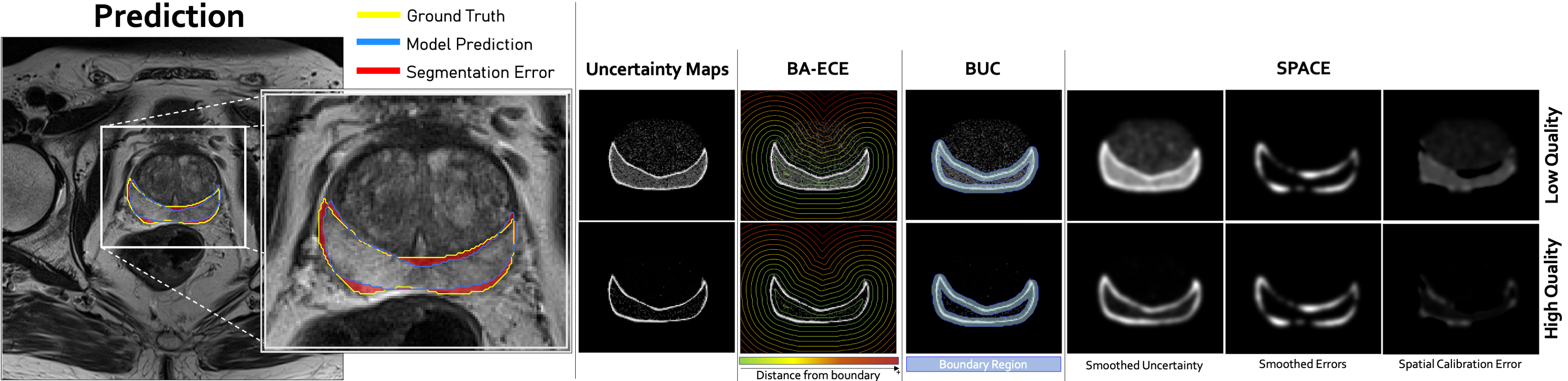}
    \caption{\textbf{Illustration of spatially aware metrics on low-quality ("noisy") and high-quality ("clean") uncertainty maps.} On the left are model predictions for segmenting the prostate peripheral zone on bi-parametric MRI (yellow contours indicate ground truth, blue represents model predictions, and red regions denote model errors). Next to that are two Monte Carlo dropout uncertainty maps: a low-quality map (top) and a high-quality map (bottom). To the right, we visualize the spatially driven metrics: the contours for spatial distance binning used in Boundary-Aware Expected Calibration Error (BA-ECE), the boundary region (light blue) used in the Boundary Uncertainty Concentration (BUC) metric (middle), and, on the right, the smoothed uncertainty and prediction error maps along with their difference, which produces the Spatial Calibration Error map for the Spatially-Aware Calibration Error (SPACE) metric. }
    \label{fig:metrics_illustration}
\end{figure*}

\section{Experiments}

\subsection{Setup and Data}
We evaluated our spatially-aware metrics on a multi-parametric prostate MRI dataset consisting of 36 cases, with ground-truth annotations delineating the peripheral and transitional zones~\cite{antonelli2022medical}. Segmentations were generated using the top-performing pre-trained nnU-Net model from the Medical Segmentation Decathlon~\cite{isensee2021nnu}. 

\subsection{Uncertainty Map Generation}
Uncertainty maps were generated using Monte Carlo dropout \cite{gal2016dropout} with \( n = 30 \) forward-pass repetitions. Dropout layers were placed after every encoder and decoder step, except for the output step. We generated two uncertainty maps per case (Figure \ref{fig:metrics_illustration}):
\begin{itemize}
    \item \textbf{Low quality map}: High fixed dropout rate producing diffused uncertainty.
    \item \textbf{High quality map}: Adaptive dropout rates (Rate-In) \cite{Zeevi_2025_CVPR} emphasizing boundary uncertainty.
\end{itemize}
To perform a fair comparison, both maps were evaluated against the same binary error map $\mathbf{E}$ produced by the full (no dropout) model.

\subsection{Analysis Protocol}
For each case, traditional and spatially-aware metrics were computed separately for the low and high quality uncertainty maps. Discriminative power was assessed through:
\begin{itemize}
    \item \textbf{Accuracy}: Percentage of cases where a metric correctly favored the high quality map.
    \item \textbf{Effect size}: Cohen’s \(d\), measuring how well a metric differentiates clean from noisy maps by standardizing their difference relative to variability across cases. Larger values indicate a greater ability to favor structured uncertainty over scattered noise.
    \item \textbf{Relative improvement}: Percentage change in metric values, computed relative to the noisy map.

\end{itemize}
Statistical significance was tested via Cochran’s Q and pairwise McNemar tests.

\subsection{Results}
We evaluated uncertainty metrics on prostate zonal segmentation and observed clear differences in their ability to distinguish high-quality ("clean") from low-quality ("noisy") uncertainty maps (Table \ref{tab:metric-comparison}, Figure \ref{fig:metrics_illustration}).

\begin{table}[h]
\caption{\textbf{Spatially-Aware Metrics Significantly Outperform Traditional Uncertainty Estimation Metrics for Segmentation Uncertainty Quality Assessment.} 
Comparison of uncertainty estimation metrics for assessing segmentation uncertainty maps. Spatially-aware metrics (shown in bold) demonstrate higher accuracy in identifying clean uncertainty maps over noisy ones. Cohen’s \textit{d} measures effect size (higher values indicate stronger discrimination), and Mean Difference quantifies the average percentage improvement of clean over noisy methods. Statistical significance is assessed via McNemar’s tests with Holm correction, comparing each focus metric against traditional metrics.}

\label{tab:metric-comparison}
\centering
\footnotesize
\setlength{\tabcolsep}{3.5pt}
\begin{tabular}{@{}p{3.15cm}ccc@{}}
\toprule
\multirow{1}{*}{Metric} & Accuracy (\%) & Cohen's \textit{d} & Mean Diff. (\%) \\
\midrule
\textbf{SPACE}$^{\S}$ & \textbf{95.83} & 1.34 & 49.8 \\
\textbf{BUC}$^{\dagger}$ & \textbf{91.67} & 1.26 & 34.1 \\
\textbf{BA-ECE}$^{\dagger}$ & \textbf{89.58} & \textbf{1.83} & \textbf{55.8} \\
\midrule
ECE$^{\text{\cite{naeini2015obtaining}}}$ & 83.33 & 1.09 & 37.7 \\
VOXEL\_ACC$^{\text{\cite{mehrtash2020confidence}}}$ & 79.17 & 0.23 & 0.1 \\
AUC-PR$^{\text{\cite{hendrycks2016baseline}}}$ & 75.00 & 0.48 & 18.1 \\
AUC-ROC$^{\text{\cite{hendrycks2016baseline}}}$ & 72.92 & 0.27 & 1.0 \\
PAvPU$_{\omega=5}$$^{\text{\cite{mukhoti2018evaluating}}}$ & 68.75 & 0.17 & 0.1 \\
AU-ARC$^{\text{\cite{nadeem2009accuracy}}}$ & 66.67 & 0.16 & 0.0  \\
MCE$^{\text{\cite{naeini2015obtaining}}}$ & 56.25 & 0.12 & 1.4 \\
PAvPU$_{\omega=11}$$^{\text{\cite{mukhoti2018evaluating}}}$ & 33.33 & 0.14 & 0.0 \\
Correct Certain Ratio$^{\text{\cite{mobiny2021dropconnect}}}$ & 31.25 & -0.10 & 0.0 \\
Uncertain Incorrect Ratio$^{\text{\cite{mobiny2021dropconnect}}}$ & 31.25 & -0.32 & -8.9 \\
\bottomrule
\end{tabular}

\begin{flushleft}
\footnotesize
Significance based on pairwise McNemar tests with Holm correction:\\
$\S$ : significantly outperforms all traditional metrics (p $\leq$ 0.05)\\
$\dagger$ : significantly outperforms metrics with $\leq$ 70\% accuracy (p $\leq$ 0.05)
\end{flushleft}
\end{table}

\vspace{-10pt}

\noindent Among the proposed spatially-aware metrics, SPACE achieved the highest accuracy (95.83\%), followed by BUC (91.67\%) and BA-ECE (89.58\%). These metrics showed stronger discrimination between high and low quality uncertainty maps compared to traditional voxel-wise metrics.

Effect size (Cohen’s d) and mean difference analyses further confirmed these findings. BA-ECE had the highest effect size (1.83) and mean difference (55.8\%), indicating a substantial separation between high quality and low quality maps. The boundary-aware extension of ECE (BA-ECE) improved over standard ECE, showing +6.25\% higher accuracy, +68\% increase in effect size, and +48\% increase in mean difference.

Pairwise McNemar tests with Holm correction confirmed that the differences between spatially-aware and voxel-wise metrics were statistically significant for SPACE. These results indicate that incorporating spatial information improves the assessment of uncertainty maps in prostate segmentation.

\section{Summary}

We propose three spatially-aware metrics for evaluating segmentation uncertainty, each highlighting a unique aspect of spatial structure:

\begin{enumerate}
\item \textbf{Boundary Uncertainty Concentration (BUC)} measures whether high uncertainty appears along anatomical edges, reflecting clinically critical boundary regions.
\item \textbf{Boundary-Aware ECE (BA-ECE)} extends calibration error by grouping predictions according to their distance from the ground-truth boundary, focusing on calibration quality where errors often matter most.
\item \textbf{Spatially-Aware Calibration Error (SPACE)} assesses local overlap between predicted uncertainty and actual errors via spatial smoothing, revealing how consistently the model flags potential mispredictions.
\end{enumerate}

Our evaluation showed that these spatially-aware metrics more effectively distinguish meaningful uncertainty patterns from scattered noise than traditional voxel-wise methods. SPACE achieved the highest accuracy and effect size, while BUC and BA-ECE offered complementary insights into boundary-aware uncertainty concentration and calibration. Notably, BA-ECE improved over standard ECE, underscoring the value of incorporating spatial structure into uncertainty evaluation.

In clinical applications, these metrics could improve segmentation reliability assessment for tasks such as automatic quality control, uncertainty-aware model selection, and clinician decision support. Since boundary regions are crucial for surgical planning and disease monitoring, better capturing uncertainty at these edges could enhance the trustworthiness of AI-driven segmentation.

While our results suggest clear advantages of spatially-aware evaluation, further exploration is needed to assess generalizability across different datasets and imaging modalities. Future work may examine these metrics on other anatomical structures and investigate integrating them into model training to encourage clinically meaningful uncertainty patterns.

{
    \small
    \bibliographystyle{ieeenat_fullname}
    \bibliography{main}

\begin{thebibliography}{16}
\providecommand{\natexlab}[1]{#1}
\providecommand{\url}[1]{\texttt{#1}}
\expandafter\ifx\csname urlstyle\endcsname\relax
  \providecommand{\doi}[1]{doi: #1}\else
  \providecommand{\doi}{doi: \begingroup \urlstyle{rm}\Url}\fi

\bibitem[Antonelli et~al.(2022)Antonelli, Reinke, Bakas, Farahani, Kopp-Schneider, Landman, Litjens, Menze, Ronneberger, Summers, et~al.]{antonelli2022medical}
Michela Antonelli, Annika Reinke, Spyridon Bakas, Keyvan Farahani, Annette Kopp-Schneider, Bennett~A Landman, Geert Litjens, Bjoern Menze, Olaf Ronneberger, Ronald~M Summers, et~al.
\newblock The medical segmentation decathlon.
\newblock \emph{Nature communications}, 13\penalty0 (1):\penalty0 4128, 2022.

\bibitem[Gal and Ghahramani(2016)]{gal2016dropout}
Yarin Gal and Zoubin Ghahramani.
\newblock Dropout as a bayesian approximation: Representing model uncertainty in deep learning.
\newblock In \emph{international conference on machine learning}, pages 1050--1059. PMLR, 2016.

\bibitem[Geifman et~al.(2018)Geifman, Uziel, and El-Yaniv]{geifman2018bias}
Yonatan Geifman, Guy Uziel, and Ran El-Yaniv.
\newblock Bias-reduced uncertainty estimation for deep neural classifiers.
\newblock \emph{arXiv preprint arXiv:1805.08206}, 2018.

\bibitem[Guo et~al.(2017)Guo, Pleiss, Sun, and Weinberger]{guo2017calibration}
Chuan Guo, Geoff Pleiss, Yu Sun, and Kilian~Q Weinberger.
\newblock On calibration of modern neural networks.
\newblock In \emph{International conference on machine learning}, pages 1321--1330. PMLR, 2017.

\bibitem[Hendrycks and Gimpel(2016)]{hendrycks2016baseline}
Dan Hendrycks and Kevin Gimpel.
\newblock A baseline for detecting misclassified and out-of-distribution examples in neural networks.
\newblock \emph{arXiv preprint arXiv:1610.02136}, 2016.

\bibitem[Isensee et~al.(2021)Isensee, Jaeger, Kohl, Petersen, and Maier-Hein]{isensee2021nnu}
Fabian Isensee, Paul~F Jaeger, Simon~AA Kohl, Jens Petersen, and Klaus~H Maier-Hein.
\newblock nnu-net: a self-configuring method for deep learning-based biomedical image segmentation.
\newblock \emph{Nature methods}, 18\penalty0 (2):\penalty0 203--211, 2021.

\bibitem[Krishnan and Tickoo(2020)]{krishnan2020improving}
Ranganath Krishnan and Omesh Tickoo.
\newblock Improving model calibration with accuracy versus uncertainty optimization.
\newblock \emph{Advances in Neural Information Processing Systems}, 33:\penalty0 18237--18248, 2020.

\bibitem[Laves et~al.(2020)Laves, Ihler, Fast, Kahrs, and Ortmaier]{laves2020well}
Max-Heinrich Laves, Sontje Ihler, Jacob~F Fast, L{\"u}der~A Kahrs, and Tobias Ortmaier.
\newblock Well-calibrated regression uncertainty in medical imaging with deep learning.
\newblock In \emph{Medical imaging with deep learning}, pages 393--412. PMLR, 2020.

\bibitem[Mehrtash et~al.(2020)Mehrtash, Wells, Tempany, Abolmaesumi, and Kapur]{mehrtash2020confidence}
Alireza Mehrtash, William~M Wells, Clare~M Tempany, Purang Abolmaesumi, and Tina Kapur.
\newblock Confidence calibration and predictive uncertainty estimation for deep medical image segmentation.
\newblock \emph{IEEE transactions on medical imaging}, 39\penalty0 (12):\penalty0 3868--3878, 2020.

\bibitem[Mobiny et~al.(2021)Mobiny, Yuan, Moulik, Garg, Wu, and Van~Nguyen]{mobiny2021dropconnect}
Aryan Mobiny, Pengyu Yuan, Supratik~K Moulik, Naveen Garg, Carol~C Wu, and Hien Van~Nguyen.
\newblock Dropconnect is effective in modeling uncertainty of bayesian deep networks.
\newblock \emph{Scientific reports}, 11\penalty0 (1):\penalty0 5458, 2021.

\bibitem[Mukhoti and Gal(2018)]{mukhoti2018evaluating}
Jishnu Mukhoti and Yarin Gal.
\newblock Evaluating bayesian deep learning methods for semantic segmentation.
\newblock \emph{arXiv preprint arXiv:1811.12709}, 2018.

\bibitem[Nadeem et~al.(2009)Nadeem, Zucker, and Hanczar]{nadeem2009accuracy}
Malik Sajjad~Ahmed Nadeem, Jean-Daniel Zucker, and Blaise Hanczar.
\newblock Accuracy-rejection curves (arcs) for comparing classification methods with a reject option.
\newblock In \emph{Machine Learning in Systems Biology}, pages 65--81. PMLR, 2009.

\bibitem[Naeini et~al.(2015)Naeini, Cooper, and Hauskrecht]{naeini2015obtaining}
Mahdi~Pakdaman Naeini, Gregory Cooper, and Milos Hauskrecht.
\newblock Obtaining well calibrated probabilities using bayesian binning.
\newblock In \emph{Proceedings of the AAAI conference on artificial intelligence}, 2015.

\bibitem[Nixon et~al.(2019)Nixon, Dusenberry, Zhang, Jerfel, and Tran]{nixon2019measuring}
Jeremy Nixon, Michael~W Dusenberry, Linchuan Zhang, Ghassen Jerfel, and Dustin Tran.
\newblock Measuring calibration in deep learning.
\newblock In \emph{CVPR workshops}, 2019.

\bibitem[Zeevi et~al.(2024)Zeevi, Venkataraman, Staib, and Onofrey]{zeevi2024monte}
Tal Zeevi, Rajesh Venkataraman, Lawrence~H Staib, and John~A Onofrey.
\newblock Monte-carlo frequency dropout for predictive uncertainty estimation in deep learning.
\newblock In \emph{2024 IEEE International Symposium on Biomedical Imaging (ISBI)}, pages 1--5. IEEE, 2024.

\bibitem[Zeevi et~al.(2025)Zeevi, Shwartz-Ziv, LeCun, Staib, and Onofrey]{Zeevi_2025_CVPR}
Tal Zeevi, Ravid Shwartz-Ziv, Yann LeCun, Lawrence~H. Staib, and John~A. Onofrey.
\newblock Rate-in: Information-driven adaptive dropout rates for improved inference-time uncertainty estimation.
\newblock In \emph{Proceedings of the Computer Vision and Pattern Recognition Conference (CVPR)}, pages 20757--20766, 2025.

\end{thebibliography}
}

\end{document}